\documentclass{nsr}

\usepackage{amsmath,graphicx,array}
\usepackage{dcolumn,soul}%

\usepackage{amsthm}
\usepackage[figuresright]{rotating}%
\usepackage{algorithm, algorithmicx, algpseudocode}
\usepackage{listings}%
\usepackage{hyperref}
\usepackage[T1]{fontenc}
\usepackage[utf8]{inputenc}
\usepackage{babel}
\usepackage{pifont}
\usepackage{pmboxdraw}
\usepackage{amsmath}

\usepackage{amssymb}
\usepackage{graphicx}
\usepackage{wasysym}
\usepackage{booktabs}
\usepackage{multirow}
\usepackage{amsthm}

\usepackage{makecell}
\usepackage{graphicx}
\usepackage{lipsum} 
\usepackage{wrapfig}
\usepackage{adjustbox}
\usepackage{lipsum} 
\providecommand{\tabularnewline}{\\}

\makeatletter
\def\uns{\ifmmode\,\else$\,$\fi}%

\makeatother
 
\jvol{XX}
\jnum{X}
\jyear{Year}
\doi{10.1093/nsr/XXXX}
\received{XX XX Year}
\revised{XX XX Year}
\accepted{XX XX Year}

\begin{document}

\dhead{RESEARCH ARTICLE}

\subhead{INFORMATION SCIENCE}

\title{Quantum-Inspired Analysis of Neural Network Vulnerabilities: The Role of Conjugate Variables in System Attacks}

\author{Jun-Jie Zhang$^{1}$}

\author{Deyu Meng$^{2,*}$}

\affil{$^1$Division of Computational physics and Intelligent modeling, Northwest
Institute of Nuclear Technology, Shaanxi, Xi’an 710024, China\\
E-mail: zjacob@mail.ustc.edu.cn}

\affil{$^2$School of Mathematics and Statistics and Ministry of Education Key Lab of Intelligent Networks and Network Security,
Xi'an Jiaotong University, Shaanxi, P. R. China.\\
Email: dymeng@mail.xjtu.edu.cn}

\authornote{\textbf{Corresponding author.} Email: dymeng@mail.xjtu.edu.cn}

\abstract[ABSTRACT]{Neural networks demonstrate inherent vulnerability to small, non-random perturbations, emerging as adversarial attacks. Such attacks, born from the gradient of the loss function relative to the input, are discerned as input conjugates, revealing a systemic fragility within the network structure. Intriguingly, a mathematical congruence manifests between this mechanism and the quantum physics' uncertainty principle, casting light on a hitherto unanticipated interdisciplinarity. This inherent susceptibility within neural network systems is generally intrinsic, highlighting not only the innate vulnerability of these networks but also suggesting potential advancements in the interdisciplinary  area for understanding these black-box networks.}



\maketitle

\section{Introduction}\label{sec1}

Despite the widely demonstrated success across various domains — from image classification\cite{10.1145/3065386} and speech recognition\cite{6296526} to predicting protein structures\cite{Senior2020}, playing chess\cite{Silver2016} and other games\cite{Schrittwieser2020}, etc. — deep neural networks have recently come under scrutiny for an intriguing vulnerability\cite{759851e20d2e47aaad2a560211f6a126,REN2020346}. The robustness of these intricately trained models is being called into question, as they seem to falter under attacks that are virtually imperceptible to human senses.

A growing body of both empirical\cite{10.1007/978-3-030-01258-8_39, 8578273,jia-liang-2017-adversarial,chen2018attacking,carlini2018audio,xu2012sparse,pmlr-v148-benz21a,morcos2018on,NEURIPS2021_50f3f8c4} and theoretical\cite{pmlr-v97-zhang19p,goodfellow2015explaining,tsipras2019robustness,antun2021can} evidences suggests that these sophisticated networks can be tripped up by minor, non-random perturbations, producing high-confidence yet erroneous predictions — a striking and quite succinct example being the Fast Gradient Sign Method (FGSM) attack\cite{goodfellow2015explaining}. These findings raise significant concerns about the vulnerabilities of such neural networks. If their performance can indeed be undermined by such slight disruptions, the reliability of technologies that hinge on state-of-the-art deep learning could potentially be at risk.

A natural question emerges concerning the vulnerability of deep neural networks. Despite the classical approximation theorems\cite{cybendo1992approximations,HORNIK1989359,10.1162/neco.1989.1.4.502,737488} promising that a neural network can approximate a continuous function to any desired level of accuracy, is the observed trade-off between accuracy and robustness an intrinsic and universal property of these networks?

This query stems from the intuition that stable problems, described by stable functions, should intrinsically produce stable solutions. The debate within the scientific community is still ongoing. If this trade-off is indeed an inherent feature, then a comprehensive exploration into the foundations of deep learning is warranted. Alternatively, if this phenomenon is merely an outcome of approaches to constructing and training neural networks, it would be beneficial to concentrate on enhancing these processes, as have already been undertaken, e.g., the certified Adversarial Robustness via Randomized Smoothing\cite{pmlr-v119-yang20c,pmlr-v162-hao22c,pmlr-v97-cohen19c}, and the concurrent training
strategy\cite{NEURIPS2020_61d77652,arani2020adversarial,arcaini2021roby,tsipras2019robustness,sehwag2021improving,leino2021globally-robust,antun2020on,rozsa2016are}, etc.

In this study, we uncover an intrinsic characteristic of neural networks: their vulnerability shares a mathematical equivalence with the uncertainty principle in quantum physics\cite{Heisenberg1927, Bohr1950}. This is observed when gradient-based attacks\cite{goodfellow2015explaining,kurakin2017adversarial,madry2018towards,7467366,7780651,8954332} on the inputs are identified as conjugate variables, in relation to these inputs.

\begin{figure}[t!]
\begin{adjustbox}{width=24pc,right}
\includegraphics{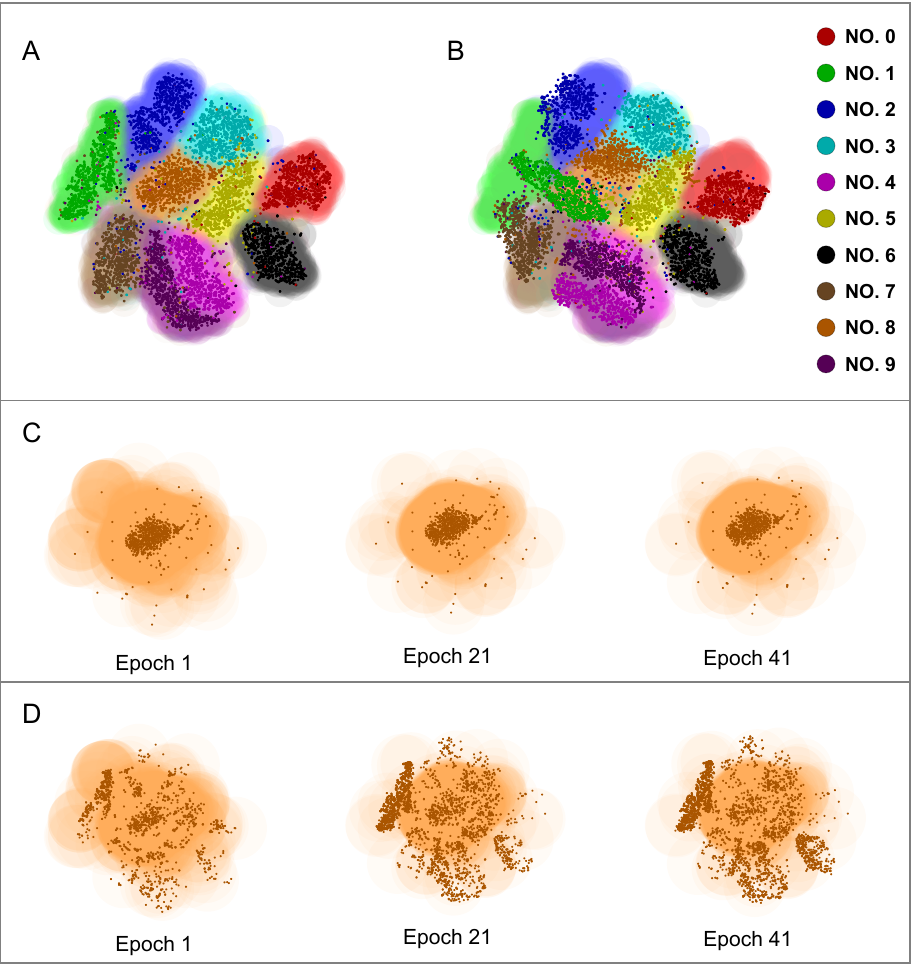} 
\end{adjustbox}
\caption{Illustration of $\Delta x$ and $\Delta p$ in a three-layer convolutional neural network trained on the MNIST dataset over 50 epochs. The data's high-dimensional feature space was reduced to two dimensions using the t-SNE (t-Distributed Stochastic Neighbor Embedding) algorithm for easy visualization. (A) Shaded regions indicate the  class predictions obtained by the finally trained network, and the colors imposed on individual points indicate the true labels of corresponding test samples. (B) All test samples were subjected to the Projected Gradient Descent (PDG) adversarial attack method \cite{kurakin2017adversarial,madry2018towards} with $\epsilon=0.1$ and $\alpha=0.1/4$ over four iterative steps. It is seen that these adversarially perturbed samples are evidently deviated from class regions they should be located. (C) The prediction region evolution for the digit '8' is displayed at epochs 1, 21, and 41. More deeper the color is, more confident the prediction is by the network. 
(D) The shaded area is similar to (C), but with points representing the adversarial predictions of the attacked images, illustrating the temporal impact of the PDG attack on model accuracy.}
\label{dx_dp_visualization}
\end{figure}

Taking into account a trained neural network model, denoted as $f(X,\theta)$, where $\theta$ signifies the parameters and $X$ represents the input variable of the network, we observe a consistent pattern. The network cannot achieve arbitrary levels of measuring certainties on two factors simultaneously: the conjugate variable $\nabla_{X}l(f(X,\theta),Y)$ (where $Y$ denotes the underlying groundtruth label of $X$) and the input $X$, leading to the observed accuracy-robustness trade-off. This phenomenon, similar to the quantum physics' uncertainty principle, offers a nuanced understanding of the limitations inherent in neural networks. 

\section{Results}\label{sec2}

\subsection{Conjugate variables as attacks}

In quantum mechanics, the concept of conjugate variables plays a critical role in understanding the fundamentals of particle behavior. Conjugate variables are a pair of observables, typically represented by operators, which do not commute. This non-commutativity implies that the order of their operations is significant and it is intrinsically tied to Heisenberg's uncertainty principle\cite{Heisenberg1927, Bohr1950}. A prime example of such a pair is the position operator, $\hat{x}_\text{qt}$, and the momentum operator, $\hat{p}_\text{qt}=-i\frac{\partial}{\partial x_\text{qt}}$. Here, the order of operations matters such that $\hat{x}_\text{qt}\hat{p}_\text{qt}$ is not equal to $\hat{p}_\text{qt}\hat{x}_\text{qt}$, indicating the impossibility of simultaneously determining the precise values of both position and momentum. This inherent uncertainty is quantitatively expressed in Heisenberg's uncertainty relation: $\Delta x_\text{qt} \Delta p_\text{qt} \geq \frac{1}{2}$, where $\Delta x_\text{qt}$ and $\Delta p_\text{qt}$ represent the standard deviations of position and momentum measurements, respectively. 

Drawing an analogy from quantum mechanics, we can formulate the concepts of conjugate variables within the realm of neural networks. Specifically, the features of the input data provided to a neural network can be conceptualized as feature operators, denoted as $\hat{x}_{i}$, while the gradients of the loss function with respect to these inputs can be viewed as attack operators, denoted as $\hat{p}_{i}=\frac{\partial}{\partial x_{i}}$. Here, the subscript $i$ refers to the $i$-th feature of the entire input feature vector. The attack operators, corresponding to the gradients on inputs, hold a clear relationship with gradient-based attacks, such as the FGSM attack (the application of such attacks often involves a sign function, although this is not strictly necessary\cite{10.1007/978-3-031-06767-9_17,ijcai2021p635}).

This analogy leads us to an inherent uncertainty relation for neural networks, mirroring the Heisenberg's uncertainty principle in quantum mechanics. Providing a trained neural network with properly normalized loss functions, the relation reads: $\Delta x_{i} \Delta p_{i} \geq \frac{1}{2}$ (see derivations in Methods). This relation, relying on both the dataset and the network structure, suggests that there exists an intrinsic limitation in precisely measuring both features and attacks simultaneously. This intrinsically reveals an inherent vulnerability of neural networks, echoing the uncertainty we observe in the quantum world.

To intuitively visualize the manifestation of $\Delta x = (\sum{\Delta x_{i}})^{1/2}$ and $\Delta p = (\sum{\Delta p_{i}})^{1/2}$ within neural networks, we use the MNIST dataset as a representative example. The neural network is trained and subsequently subjected to attacks at each training epoch.

In this scenario, a trained network partitions the hyperspace (the space inhabited by the samples) into distinct regions. A given input, represented as a point in this space, is classified based on the label of the region it falls within. After 50 epochs of training, the shaded areas encapsulate most correctly labeled data points (Fig. \ref{dx_dp_visualization}A). Conversely, the attacks shift these input points slightly, leading to misclassification. The shifted points do not overlap with the regions defined by the trained network (Fig. \ref{dx_dp_visualization}B).

We pay particular attention to class number 8, which exhibits the most interconnections with other classes. This class is further illustrated in Fig. \ref{dx_dp_visualization}C and D. As the training epochs progress, the "effective radius" of the shaded area shrinks, causing the area to gradually coincide with the correctly labeled data points (Fig. \ref{dx_dp_visualization}C). Simultaneously, the "effective radius" of the attacked points begins to deviate further from the shaded regions, and thus from the correctly labeled data (Fig. \ref{dx_dp_visualization}D).

This visualization reveals an inherent trade-off: a reduction in the effective radius of the trained class corresponds to an increase in the effective radius of the attacked points. These two radii can be conceptualized as the visual representations of the uncertainties, $\Delta x$ and $\Delta p$, highlighting the delicate balance of precision and vulnerability in neural networks.

In addition to the adversarial attacks explored in this study, there exist analogous effective conjugates in other types of adversarial attacks as well\cite{kurakin2017adversarial,7467366,7780651,8954332}. While we are currently unable to explicitly define the conjugates associated with black-box attacks as referenced in \cite{8601309, 10.1007/978-3-030-58592-1_29}, it is plausible that these methods may adhere to the same underlying principle.

\subsection{Manifestation of the uncertainty principle in neural networks}

\begin{figure*}[th!]
\begin{adjustbox}{width=33pc, center}
\includegraphics{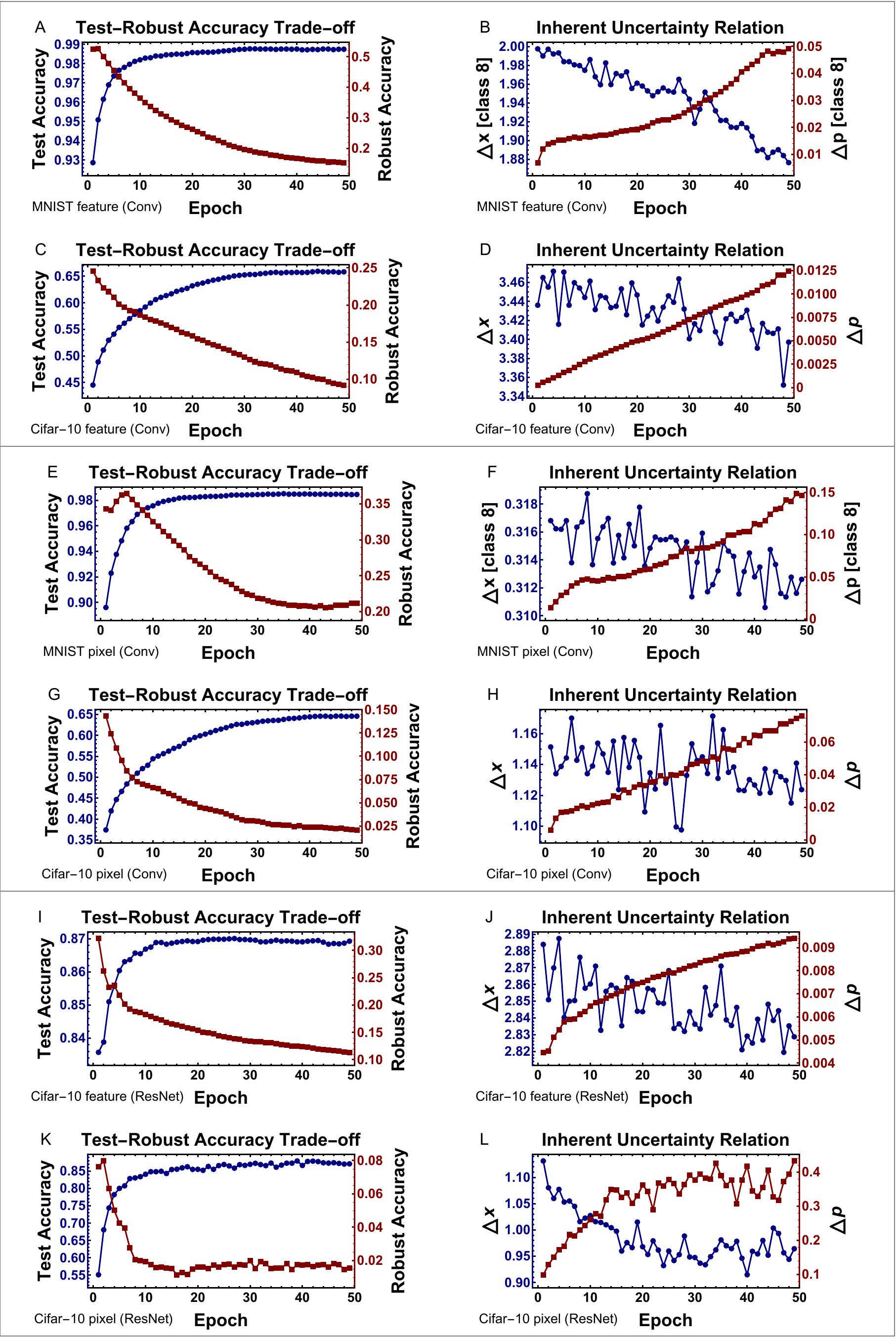} 
\end{adjustbox}
\caption{Results of the three different types of neural networks: a three-layer convolutional network running on the MNIST dataset, a four-layer convolutional network on the CIFAR-10 dataset, and a residual network\cite{7780459} with eight convolutional layers on the CIFAR-10 dataset. The term "feature" in the labels represents the results obtained by attacking the features of the input images, while "pixel" corresponds to attacks directed at the pixels themselves. Each neural network underwent training for a span of 50 epochs. The quantities $\Delta x$ and $\Delta p$ were determined through high-dimensional Monte-Carlo integrations. Subfigures (A), (C), (E), (G), (I), and (K) depict the test and robust accuracy metrics, with the robust accuracy evaluated on images perturbed by the PDG adversarial attack method, using parameters $\epsilon=8/255$ and $\alpha=2/255$ across four iterative steps. Subfigures (B), (D), (F), (H), (J), and (L) illustrate the trade-off relationship between $\Delta x$ and $\Delta p$.}
\label{TA_RA}
\end{figure*}

The shaded areas in Fig. \ref{dx_dp_visualization}A are actually representative of wave functions in quantum physics. Specifically, for the MNIST dataset, we have ten corresponding wave functions corresponding to ten digit number classes. Therefore, the  uncertainty relation $\Delta x \Delta p \geq \frac{1}{2}$ shown in Fig. \ref{dx_dp_visualization}C and D should be reinterpreted as $\Delta x[\text{class 8}] \Delta p[\text{class 8}] \geq \frac{1}{2}$, indicating that we are concentrating on the class of number 8. This equation is a clear depiction of the trade-off between $\Delta x[\text{class 8}]$ and $\Delta p[\text{class 8}]$, as depicted in Fig. \ref{TA_RA}B, accompanied by the associated trade-off between accuracy and robustness (Fig. \ref{TA_RA}A). The Cifar-10 dataset, having a higher complexity than MNIST, poses a potential indeterminacy in identifying a specific class that has more connectivity with other classes. In this case, the average values $\Delta x = \text{Mean}(\Delta x[\text{All classes}])$ and $\Delta p = \text{Mean}(\Delta p[\text{All classes}])$ are employed instead. The similar results obtained on Cifar-10 underscore the inherent uncertainty relation that drives the accuracy-robustness trade-off, as demonstrated in Fig. \ref{TA_RA}C and D. 

\section{Discussions}\label{sec3}

\begin{table*}[t]
\centering
\caption{Comparison of the uncertainty principle between quantum physics and neural networks.\label{tab:Comparison-of-the uncertainty} The subscript $i$ represents the $i$-th dimension. For physics, $i$ stands for the spatial coordinates ($x$, $y$, and $z$), whereas in the context of neural networks, $i$ refers to the $i$-th feature. When we consider pixels, $i$ simply pertains to the $i$-th pixel. Additionally, we utilize Dirac notation, for instance, $\langle\hat{x}_{i,\text{qt}}\rangle = \int\psi^{*}(X)x_{i,\text{qt}}\psi(X)dX$, where $\langle\hat{x}_{i,\text{qt}}\rangle$ is the expectation value of the $i$-th dimension. Similarly, $\langle\hat{x}_{i}\rangle = \int\psi_{Y}(X)x_{i}\psi_{Y}(X)dX$ for neural networks.}
\resizebox{\linewidth}{!}{
\begin{tabular}{cc|cc}
\toprule
\multicolumn{2}{c}{Quantum physics} & \multicolumn{2}{c}{Neural networks}\tabularnewline
\midrule
position & $X=(x,y,z)$ & $X=(x_{1},...,x_{i},...,x_{M})$ & image/feature (input)\tabularnewline
\midrule
\begin{tabular}[c]{@{}c@{}}momentum \\ (conjugate of position)\end{tabular} & $P=(p_{x},p_{y},p_{z})$ & $P=(p_{1},...,p_{i},...,p_{M})$ & \begin{tabular}[c]{@{}c@{}}attack \\(conjugate of input)\end{tabular}\tabularnewline
\midrule
wave function & $\psi(X)$ & $\psi_{Y}(X)$ & \begin{tabular}[c]{@{}c@{}}normalized loss function \\(neural packet)\end{tabular}\tabularnewline
\midrule
normalize condition & $\int|\psi(X)|^{2}dX=1$ & $\int|\psi_{Y}(X)|^{2}=1$ & normalize condition\tabularnewline
\midrule
position operator & $\hat{x}_{i,\text{qt}}\psi(X)=x_{i,\text{qt}}\psi(X)$ & $\hat{x}_{i}\psi_{Y}(X)=x_{i}\psi_{Y}(X)$ & feature operator\tabularnewline
\midrule
momentum operator & $\hat{p}_{i,\text{qt}}\psi(X)=-i\frac{\partial}{\partial x_{i,\text{qt}}}\psi(X)$ & $\hat{p}_{i}\psi_{Y}(X)=\frac{\partial}{\partial x_{i}}\psi_{Y}(X)$ & attack operator\tabularnewline
\midrule
\begin{tabular}[c]{@{}c@{}}standard deviation for \\ measuring position\end{tabular} & $\sigma_{x_{i,\text{qt}}}=\langle(\hat{x}_{i,\text{qt}}-\langle\hat{x}_{i,\text{qt}}\rangle)^{2}\rangle^{1/2}$ & $\Delta{x_{i}}=\langle(\hat{x}_{i}-\langle\hat{x}_{i}\rangle)^{2}\rangle^{1/2}$ & \begin{tabular}[c]{@{}c@{}}standard deviation for \\ resolving pixel\end{tabular}\tabularnewline
\midrule
\begin{tabular}[c]{@{}c@{}}standard deviation for \\ measuring momentum\end{tabular} & $\sigma_{p_{i,\text{qt}}}=\langle(\hat{p}_{i,\text{qt}}-\langle\hat{p}_{i,\text{qt}}\rangle)^{2}\rangle^{1/2}$ & $\Delta{p_{i}}=\langle(\hat{p}_{i}-\langle\hat{p}_{i}\rangle)^{2}\rangle^{1/2}$ & \begin{tabular}[c]{@{}c@{}}standard deviation for \\ resolving attack\end{tabular}\tabularnewline
\midrule
uncertainty relation & $\sigma_{x_{i,\text{qt}}}\sigma_{p_{i,\text{qt}}}\geq\frac{1}{2}$ & $\Delta{x_{i}}\Delta{p_{i}}\geq\frac{1}{2}$ & uncertainty relation\tabularnewline
\bottomrule
\end{tabular}}
\end{table*}

\subsection{Attacking features is more effective than attacking pixels}

The pixels in our dataset serve as the raw, unprocessed data, gathered directly from the detectors. These pixels carry the features that serve as an accurate representation of the real world. While there is a possibility of manipulating these features, it is more common and practical to focus on the pixels themselves. By doing so, we can observe the accuracy-robustness trade-off (Fig. \ref{TA_RA}E and G), a fundamental concept that is underpinned by the uncertainty relation, as seen in Fig. \ref{TA_RA}F and H.

However, it is important to note, as evidenced by the testing accuracy results from the MNIST dataset, that there is an initial learning curve or `kick' that is encountered (Fig. \ref{TA_RA}E). This is to be expected as the neural network must first familiarize itself with, or `learn', the features before it can effectively classify the images.

While processing the initial learning stages, it is also worth noting the fluctuation in both $\Delta x$ and $\Delta p$ for input pixels. This fluctuation is more pronounced than that seen in the features, highlighting the random exploration nature of the learning algorithm. As illustrated in Fig. \ref{TA_RA}H, these fluctuations could be attributed to the inherent randomness of the learning process, a factor that is crucial to potentially uncover more optimal weight configurations.

\subsection{Phenomenon in attacking well designed neural networks}

Typically, network structures are scrupulously architected to fit the demands of specific tasks. Take Fig. \ref{TA_RA}C,D,G,H as an example. In the figure, the network only achieves a test accuracy of around 65\% due to the relatively simple network architecture. To address this, we introduce a more advanced network structure that incorporates residual networks and additional convolutional layers. This refined structure increases the accuracy to nearly 90\%\footnote{Given that the quantities $\Delta x$ and $\Delta p$ are approximately computed through high-dimensional Monte Carlo integrations, a process that is exceedingly time-consuming, we can only feasibly perform these computations for the network with such complexity. If they could be calculated more accurately under more complex and accurate networks with stronger computational resources, we believe the calculated patterns will better conform to the expected regularities.}. 
One can still observe a clear pattern in the trade-off between $\Delta x$ and $\Delta p$ for both features and pixels (Fig. \ref{TA_RA}I-L). Besides, this trade-off is also more pronounced for features than for pixels. Understanding this trade-off allows for a more effective optimization of the network structure. In closing, constructing a network structure that best fits the task at hand is pivotal in delivering optimal performance.

\subsection{Neural network as a complex physical system}

As scientific research and engineering become increasingly reliant on artificial intelligence (AI) methods, questions about the future role of human beings in these fields naturally arise. Whether guiding AI or being guided by it, understanding the fundamental principles underpinning these sophisticated structures is paramount. One approach to glean this understanding is to treat neural networks as complex physical systems, thereby applying principles of physics to elucidate the inner mechanisms of AI.

In the study at hand, it is posited that neural networks, much like quantum systems, are subject to a form of the uncertainty principle. This connection potentially uncovers intrinsic vulnerabilities within the neural networks. A comparison of formulas from these distinct fields is presented in Table \ref{tab:Comparison-of-the uncertainty}. Here, concepts from quantum physics such as position, momentum, and wave function are juxtaposed with their counterparts in neural networks: image, attack, normalized loss function, and so on. This comparison not only reveals striking similarities but also indicates that the methodologies employed in physical sciences could potentially be harnessed to investigate the properties of neural networks.

The intersection of AI and physics has the potential to provide novel insights into the intricate complexities of neural networks. For instance, the emergent capabilities exhibited by large language models might be correlated with principles found in statistical physics. Moreover, phenomena such as small data learning could be linked to concepts from Noether's theorem and gauge transformations\cite{10.1063/1.2807734}. By drawing inspiration from physical processes such as weak interactions, we can devise innovative generative models, such as "Yukawa Generative Models"\cite{Liu2023GenPhysFP}. Viewing neural networks through the lens of physics can give us a deeper understanding of their structure and functionality from an entirely new perspective. 

The synergy between AI and physics, two seemingly distinct fields, could lead to advancements in both domains. It's a two-fold benefit: AI could gain from the structured, universal laws of physics, and in return, physics could possibly leverage the predictive and analytical power of AI.

\section{Conclusion}\label{sec4}

This study reveals the remarkable link between quantum physics and neural networks, demonstrating that these artificial systems, like quantum systems, are subject to the uncertainty principle. This principle, often associated with precision and vulnerability trade-offs, provides new insights into the potential frailties inherent in neural networks.

Our findings also indicate that attacking the features of a neural network can be more effective than focusing on its pixels. This insight could possibly influence the optimization of network structures for better performance.

Meanwhile, viewing neural networks as complex physical systems allows us to apply principles from physics to understand the behaviour of these AI systems better. This interdisciplinary approach not only enhances our comprehension of AI systems but also suggests a wealth of potential applications and advancements in both fields.

As we move forward, further exploration of this accuracy-robustness trade-off and its influence on the design of neural networks will be crucial. While this study provides a valuable perspective on the relationship between quantum physics and AI, additional research is still needed to more comprehensively understand how these principles can be applied to improve neural network robustness and design.

\section{Methods}\label{sec5}

Detailed methods and materials are given in the online supplementary data.

\section{Data Availability}\label{sec6}

All data are available in the main text or the supplementary materials. Additional data related to this paper are available at https://doi.org/10.7910/DVN/SWDL1S and 
https://doi.org/10.48550/arXiv.2205.01493.


\bibliographystyle{nsr}
\bibliography{ref}

\end{document}



\baselineskip24pt


\maketitle


\begin{sciabstract}
 In this supplementary material, we provide the detailed definitions and proofs of the uncertainty principle proposed in the main text. Meanwhile, the numerical methods for estimating the high-dimensional integrals of the parameters are also presented.
\end{sciabstract}


\section{Uncertainty Relation of the neural networks}

\subsection{Uncertainty principle in quantum physics}

In quantum physics, we can describe a particle by a wave packet $\psi(X)$
in the coordinate representation with respect to the coordinate reference
frame. The normalization condition for $\psi(X)$ is given by
\begin{eqnarray}
\int|\psi(X)|^{2}dX & = & 1,\label{eq:normalization}
\end{eqnarray}
where the square amplitude $|\psi(X)|^{2}$ gives the probability
density for finding a particle at position $X=(x,y,z)$. To measure the physical
quantities of the particle, such as position $X$ and momentum $P=(p_{x},p_{y},p_{z})$,
we need to define the position and momentum operators $\hat{x}_{i}$
and $\hat{p}_{i}$ as:
\begin{eqnarray}
\hat{x}_{i}\psi(X) & = & x_{i}\psi(X),\nonumber \\
\hat{p}_{i}\psi(X) & = & -i\frac{\partial}{\partial x_{i}}\psi(X),\label{eq:operators}
\end{eqnarray}
where $i=1,2,3$ denote the $x,y,z$ components in the coordinate
space, respectively. The
average position and momentum of the particle can be evaluated by
\begin{eqnarray}
\langle\hat{x}_{i}\rangle & = & \int\psi^{*}(X)x_{i}\psi(X)dX\nonumber \\
\langle\hat{p}_{i}\rangle & = & \int\psi^{*}(X)[-i\frac{\partial}{\partial x_{i}}\psi(X)]dX,\label{eq:average_values}
\end{eqnarray}
where $\langle\cdot\rangle$ is the Dirac symbol widely used in physics
and $\psi^{*}(X)$ is the complex conjugate of $\psi(X)$.
The standard deviations of the position $\sigma_{x_{i}}$ and momentum
$\sigma_{p_{i}}$ are defined respectively as:
\begin{eqnarray}
\sigma_{x_{i}} & = & \langle(\hat{x}_{i}-\langle\hat{x}_{i}\rangle)^{2}\rangle^{1/2},\nonumber \\
\sigma_{p_{i}} & = & \langle(\hat{p}_{i}-\langle\hat{p}_{i}\rangle)^{2}\rangle^{1/2}.\label{eq:sigma_qu}
\end{eqnarray}

In the year of 1927, Heisenberg introduced the first formulation of the
uncertainty principle in his German article\cite{Heisenberg1927}.
The Heisenberg's uncertainty principle asserts a fundamental limit
to the accuracy for certain pairs. Such variable pairs are known as complementary variables (or
canonically conjugate variables). The formal inequality relating the
standard deviation of position $\sigma_{x_{i}}$ and the standard
deviation of momentum $\sigma_{p_{i}}$ reads
\begin{eqnarray}
\sigma_{x_{i}}\sigma_{p_{i}} & \geq & \frac{1}{2}.\label{eq:Un_re_Qu}
\end{eqnarray}
Uncertainty relation Eq. (\ref{eq:Un_re_Qu}) states a fundamental
property of quantum systems and can be understood in terms of the Niels
Bohr\textquoteright s complementarity principle\cite{Bohr1950}. That is, objects have certain pairs of complementary properties cannot
be observed or measured simultaneously.

\subsection{Formulas and notations for neural networks}
Without loss of generality, we can assume
that the loss function $l(f(X,\theta),Y)$ is square integrable\footnote{In practical applications, it is rational to only consider the loss function in a limited range $l(f(X,\theta),Y)<C$ under a large constant $C$, since samples out of this range can be seen as outliers and meaningless to the problem. The loss function can then be generally guaranteed to be square integrable in this functional range.
},
\begin{eqnarray}
\int l(f(X,\theta),Y)^{2}dX & = & \beta.\label{eq:normalize_ne}
\end{eqnarray}
Eq. (\ref{eq:normalize_ne}) allows us to further normalize the loss
function as
\begin{eqnarray}
\psi_{Y}(X) & = & \frac{l(f(X,\theta),Y)}{\beta^{1/2}},\label{eq:L}
\end{eqnarray}
so that
\begin{eqnarray}
\int \psi_{Y}(X)^{2}dX=1.\label{eq:}
\end{eqnarray}
For convenience, we refer $\psi_{Y}(X)$ as a neural packet in the later discussions.
Note that under different labels $Y$, a neural network will be with a set of neural packets.

An image $X=(x_{1},...,x_{i},...,x_{M})$ with $M$ pixels can be seen
as a point in the multi-dimensional space, where the numerical values
of $(x_{1},...,x_{i},...,x_{M})$ correspond to the pixel values. The
feature and attack operators of the neural packet $\psi_{Y}(X)$ can then be
defined as:
\begin{eqnarray}
\hat{x}_{i}\psi_{Y}(X) & = & x_{i}\psi_{Y}(X),\nonumber \\
\hat{p}_{i}\psi_{Y}(X) & = & \frac{\partial}{\partial x_{i}}\psi_{Y}(X).\label{eq:operator_nu}
\end{eqnarray}
Similar as Eq. (\ref{eq:average_values}), the average pixel value at $x_{i}$ associated with neural packet $\psi_{Y}(X)$ can be evaluated as
\begin{eqnarray}
\langle\hat{x}_{i}\rangle & = & \int \psi_{Y}^*(X)x_{i}\psi_{Y}(X))dX.\label{eq:ave_x_nu}
\end{eqnarray}
Since $\psi_{Y}(X)$ corresponds to a purely real number without imaginary part, the above equation is equivalent to:
\begin{eqnarray}
\langle\hat{x}_{i}\rangle & = & \int \psi_{Y}(X)x_{i}\psi_{Y}(X))dX.\label{eq:ave_x_nu}
\end{eqnarray}

Besides, the attack operator $\hat{p}_{i}=\frac{\partial}{\partial x_{i}}$
corresponds to the conjugate variable of $x_{i}$. And we can obtain the average value for $\hat{p_{i}}$ as
\begin{eqnarray}
\langle\hat{p}_{i}\rangle & = & \int \psi_{Y}(X)\frac{\partial}{\partial x_{i}}\psi_{Y}(X)dX.\label{eq:ave_p_nu}
\end{eqnarray}

\subsection{Derivation of the uncertainty relation}\label{sec:direvation-detail}

The uncertainty principle of a trained neural network can then be deduced by the following theorem:
\begin{quote}
The standard deviations $\sigma_{{p}_{i}}$ and $\sigma_{{x}_{i}}$ corresponding to the attack and feature operators $\hat{p_{i}}$ and $\hat{x_{i}}$, respectively, are restricted by the relation:
\begin{eqnarray}
\sigma_{{p}_{i}}\sigma_{{x}_{i}} & \geq & \frac{1}{2}.
\end{eqnarray}
\end{quote}

We first introduce the standard deviations $\sigma_{a}$ and $\sigma_{b}$
corresponding to two general operators $\hat{A}$ and \textbf{$\hat{B}$}. Then it follows that:
\begin{eqnarray}
\sigma_{a}\sigma_{b} = \langle(\hat{A}-\langle\hat{A}\rangle)^{2}\rangle^{\frac{1}{2}}\langle(\hat{B}-\langle\hat{B}\rangle)^{2}\rangle^{\frac{1}{2}} \equiv \langle\hat{a}^{2}\rangle^{\frac{1}{2}}\langle\hat{b}^{2}\rangle^{\frac{1}{2}}.\nonumber \\
\label{eq:sasb}
\end{eqnarray}
In general, for any two unbounded real operators $\langle\hat{a}\rangle$
and \textbf{$\langle\hat{b}\rangle$}, the following relation holds
\begin{eqnarray}
0\le\langle(\hat{a}-i\hat{b})^{2}\rangle = \langle\hat{a}^{2}\rangle-i\langle\hat{a}\hat{b}-\hat{b}\hat{a}\rangle+\langle\hat{b}^{2}\rangle.\nonumber \\
\label{eq:prove1}
\end{eqnarray}
If we further replace $\hat{a}$ and $\hat{b}$ in Eq. (\ref{eq:prove1}) by operators $\hat{a}\langle\hat{a}^{2}\rangle^{-1/2}$ and $\hat{b}\langle\hat{b}^{2}\rangle^{-1/2}$, we can then obtain the property $2\langle\hat{a}^{2}\rangle^{1/2}\langle\hat{b}^{2}\rangle{}^{1/2} \geq i\langle\hat{a}\hat{b}-\hat{b}\hat{a}\rangle$,
which gives the basic bound for the commutator $[\hat{a},\hat{b}]\equiv\hat{a}\hat{b}-\hat{b}\hat{a}$,
\begin{eqnarray}
\langle\hat{a}^{2}\rangle^{\frac{1}{2}}\langle\hat{b}^{2}\rangle^{\frac{1}{2}} & \geq & |i\frac{1}{2}\langle[\hat{a},\hat{b}]\rangle|.\label{eq:general_uncertainty}
\end{eqnarray}
Seeing the fact that $[\hat{a},\hat{b}]=[\hat{A},\hat{B}]$, we finally obtain the uncertainty relation
\begin{eqnarray}
\sigma_{a}\sigma_{b} & \geq & |i\frac{1}{2}\langle[\hat{A},\hat{B}]\rangle|.\label{eq:general_uncertainty2}
\end{eqnarray}

In terms of the neural networks, we can simply replace operators $\hat{A}$
and $\hat{B}$ by $\hat{p}_{i}$ and $\hat{x}_{i}$ introduced in
Eq. (\ref{eq:operator_nu}), and this leads
to
\begin{eqnarray}
\sigma_{{p}_{i}}\sigma_{{x}_{i}} & \geq & |i\frac{1}{2}\langle[\hat{p}_{i},\hat{x}_{i}]\rangle| = \frac{1}{2},\label{eq:uncertainty_relation}
\end{eqnarray}
where we have used the relation
\begin{eqnarray}
[\hat{p}_{i},\hat{x}_{i}]\psi_{Y}(X) & = & [\hat{p}_{i}\hat{x}_{i}-\hat{x}_{i}\hat{p}_{i}]\psi_{Y}(X)\nonumber \\
& = & \frac{\partial}{\partial x_{i}}[x_{i}\psi_{Y}(X)] \nonumber \\
& & -x_{i}\frac{\partial}{\partial x_{i}}\psi_{Y}(X)\nonumber \\
 & = & \psi_{Y}(X).\label{eq:eigencommutator}
\end{eqnarray}

Note that for a trained neural network, $\psi_{Y}(X)$ depends on the dataset and the structure of the network. Eq. (\ref{eq:uncertainty_relation}) is a general result
for general neural networks. 

In the FGSM attack, the attacked image is of the form:
\begin{eqnarray}
X & = & X_{0}+\epsilon\cdot sign(\nabla_{X}l(f(X,\theta),Y^{*})| _{X=X_0})\nonumber \\
 & \sim & X_{0}+\epsilon\cdot\nabla_{X}l(f(X,\theta),Y^{*})| _{X=X_0}\nonumber \\
 & = & X_{0}+\epsilon\cdot\nabla_{X}[\beta^{1/2}\psi_{Y^*}(X_0)]\nonumber \\
 & = & X_{0}+\epsilon^{\prime}\hat{P}\psi_{Y^*}(X_0),\label{eq:attack_X}
\end{eqnarray}
where $\hat{P}=(\frac{\partial}{\partial x_{1}},...,\frac{\partial}{\partial x_{i}},...,\frac{\partial}{\partial x_{M}})$
and $\epsilon^{\prime}=\epsilon\cdot\beta^{1/2}$. In the second
line of Eq. (\ref{eq:attack_X}) we have used the property substantiated in \cite{agarwal2020the}: ''even
without the 'Sign' of the FGSM, a successful attack can also be achieved". From Eq. (\ref{eq:attack_X}), we can then obtain
\begin{eqnarray}
\hat{P}\psi_{Y^*}(X_0)  \sim \epsilon/\epsilon^{\prime}\cdot sign(\nabla_{X}l(f(X,\theta),Y^{*})| _{X=X_0}),\label{eq:definition_attack}
\end{eqnarray}
which is the reason that we call $\hat{p_{i}}$ the attack operator.

\section{Evaluation of  $\Delta x$ and $\Delta p$}

\subsection{Approximation of  $\Delta x$ and $\Delta p$}

In the equation referred to as Eq. (\ref{eq:sigma_qu}), we encounter complex integrals involving $\sigma_{x_{i}}$ and $\sigma_{p_{i}}$. These integrals are based on loss functions from trained neural networks and are challenging due to their high dimensionality. Specifically, they are 784-dimensional for the MNIST dataset and 3072-dimensional for the Cifar-10 dataset, which makes them impractical to calculate directly.

To work around this complexity, we simplify these multidimensional problems to a single dimension. Here's how we do it using the MNIST dataset as an example:

We start by calculating the average value of all the input pixels, which we call $X_{base}$. Then, for a given trained classifier with a loss function, $l(f(X,\theta),Y=8)$—where $Y=8$ refers to the loss associated with the label number eight—we focus on one particular dimension, $i$, of the input $X$. We keep all other dimensions fixed at their base values, $X_{base}$. This reduces the loss function to depend on just one variable, $x_{i}$.

As a result, the complex equation (Eq. (\ref{eq:normalize_ne})) simplifies to the following one-dimensional integral:

\begin{eqnarray}
\int l(f(X,\theta),Y=8)^{2}dX \Rightarrow \int l(f(x_{i},\theta),Y=8)^{2}dx_{i} = \beta_{i}(Y=8).\label{eq:int_reduce}
\end{eqnarray}

This integral can now be solved using the direct Monte-Carlo integration method. By repeating a similar process, we can calculate various quantities such as $\psi_{Y=8}(x_{i})$, $\langle\hat{x}{i}(Y=8)\rangle$, $\langle\hat{p}{i}(Y=8)\rangle$, $\sigma_{x_{i}}(Y=8)$, and $\sigma_{p_{i}}(Y=8)$.

To get an overall estimate for label number eight, we randomly pick different $i$ dimensions and then average them using the square-root of the sum:

\begin{eqnarray}
\Delta X_{8}\sim(\sum_{i}\sigma_{x_{i}}(Y=8))^{1/2},\
\Delta P_{8}\sim(\sum_{i}\sigma_{p_{i}}(Y=8))^{1/2}.\label{eq:dx_dp}
\end{eqnarray}

This approach provides only an approximate estimate of the original high-dimensional integrals. While the results may not match the exact values, this estimation is useful as long as we are interested in the comparative trend of $\Delta X$ and $\Delta P$, rather than their absolute values. Thus, this approximation is considered acceptable for our purposes.

\subsection{Integral with Respect to Features and Pixels}

\begin{figure}[th!]
\begin{adjustbox}{width=40pc, center}
\includegraphics{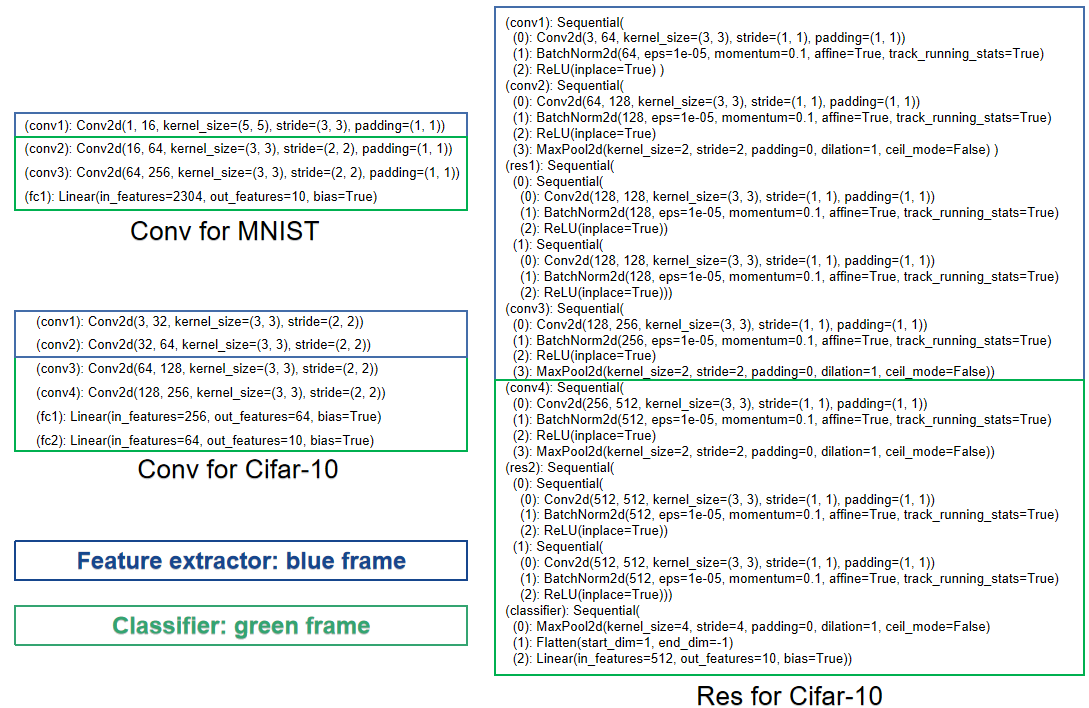} 
\end{adjustbox}
\caption{The three network structures used.}
\label{netstru}
\end{figure}

In our research, we have employed three distinct neural network architectures. Upon completion of their training, these networks are partitioned into two segments: the feature extractors and the classifiers, as illustrated in Fig. \ref{netstru}.

The integration process outlined in Equation (\ref{eq:int_reduce}) necessitates the use of an integrand space. This space can consist of either the unprocessed images at the pixel level or the attributes of the images that have undergone processing. Our study takes both scenarios into account in order to compare how the uncertainty principle manifests at each of these levels. 

When performing the integration over pixel values, the loss functions associated with the three neural networks serve as the integrands and are evaluated using the Monte Carlo technique, effectively operating within the pixel space. 

Alternatively, we initially process the images using the feature extractors, then we retrain the classifiers with randomized weights to yield three refined classifiers. The loss functions related to these classifiers are then incorporated into Equation (\ref{eq:int_reduce}) to derive the values of $\Delta X$ and $\Delta P$. In this instance, the integration is carried out over the space of extracted features.

\bibliographystyle{nsr}
\bibliography{ref}